\documentclass[letterpaper]{article} % DO NOT CHANGE THIS
\usepackage{aaai24}  % DO NOT CHANGE THIS
\usepackage{times}  % DO NOT CHANGE THIS
\usepackage{helvet}  % DO NOT CHANGE THIS
\usepackage{courier}  % DO NOT CHANGE THIS
\usepackage[hyphens]{url}  % DO NOT CHANGE THIS
\usepackage{graphicx} % DO NOT CHANGE THIS
\usepackage{natbib}  % DO NOT CHANGE THIS AND DO NOT ADD ANY OPTIONS TO IT
\usepackage{caption} % DO NOT CHANGE THIS AND DO NOT ADD ANY OPTIONS TO IT
\usepackage{algorithm}
\usepackage{algorithmic}
\usepackage{amsmath}
\usepackage{amssymb}
\usepackage[table,xcdraw]{xcolor}
\usepackage{adjustbox}
\usepackage{pifont}

\usepackage{newfloat}
\usepackage{listings}
\DeclareCaptionStyle{ruled}{labelfont=normalfont,labelsep=colon,strut=off} % DO NOT CHANGE THIS
\floatstyle{ruled}
\newfloat{listing}{tb}{lst}{}
\floatname{listing}{Listing}
\title{Disentangled Diffusion-Based 3D Human Pose Estimation with \\  Hierarchical Spatial and Temporal Denoiser}
\author{
    Qingyuan Cai\textsuperscript{\rm 1}\equalcontrib,
    Xuecai Hu\textsuperscript{\rm 1}\equalcontrib,
    Saihui Hou\textsuperscript{\rm 1,}\textsuperscript{\rm 2}\thanks{Corresponding author},
    Li Yao\textsuperscript{\rm 1},
    Yongzhen Huang\textsuperscript{\rm 1,}\textsuperscript{\rm 2}\footnotemark[2]
}

\affiliations{
    \textsuperscript{\rm 1}School of Artificial Intelligence, Beijing Normal University\\
    \textsuperscript{\rm 2}WATRIX.AI
    caiqingyuan@mail.bnu.edu.cn, \{huxc1208, housaihui, yaoli, huangyongzhen\}@bnu.edu.cn
}

\usepackage{bibentry}
\usepackage[colorlinks=true, urlcolor=blue, citecolor=black]{hyperref}
\begin{document}

\maketitle

\begin{abstract}
Recently, diffusion-based methods for monocular 3D human pose estimation have achieved state-of-the-art (SOTA) performance by directly regressing the 3D joint coordinates from the 2D pose sequence. Although some methods decompose the task into bone length and bone direction prediction based on the human anatomical skeleton to explicitly incorporate more human body prior constraints, the performance of these methods is significantly lower than that of the SOTA diffusion-based methods. This can be attributed to the tree structure of the human skeleton. Direct application of the disentangled method could amplify the accumulation of hierarchical errors, propagating through each hierarchy. Meanwhile, the hierarchical information has not been fully explored by the previous methods. To address these problems, a \textbf{D}isentangled \textbf{D}iffusion-based 3D Human \textbf{Pose} Estimation method with \textbf{H}ierarchical Spatial and Temporal Denoiser is proposed, termed \textbf{DDHPose}. In our approach: (1) We disentangle the 3D pose and diffuse the bone length and bone direction during the forward process of the diffusion model to effectively model the human pose prior. A disentanglement loss is proposed to supervise diffusion model learning. (2) For the reverse process, we propose Hierarchical Spatial and Temporal Denoiser (\textbf{HSTDenoiser}) to improve the hierarchical modeling of each joint. Our HSTDenoiser comprises two components: the Hierarchical-Related Spatial Transformer (\textbf{HRST}) and the Hierarchical-Related Temporal Transformer (\textbf{HRTT}). HRST exploits joint spatial information and the influence of the parent joint on each joint for spatial modeling, while HRTT utilizes information from both the joint and its hierarchical adjacent joints to explore the hierarchical temporal correlations among joints. Extensive experiments on the Human3.6M and MPI-INF-3DHP datasets show that our method outperforms the SOTA disentangled-based, non-disentangled based, and probabilistic approaches by 10.0\%, 2.0\%, and 1.3\%, respectively. Code and models are available at \url{https://github.com/Andyen512/DDHPose}
\end{abstract}

\section{Introduction}
3D Human Pose Estimation (HPE) has crucial applications in virtual reality ~\cite{hagbi2010shape}, human motion recognition ~\cite{zhang2022spatial}, and human-computer interaction ~\cite{kisacanin2005real}. The goal is to regress the 3D joints locations of a human in the 3D space using the input of 2D pose sequence. Most of the methods first derive predictions of 2D joints using estimators such as HRNet~\cite{wang2020deep}, CPN~\cite{chen2018cascaded}, OpenPose~\cite{cao2017realtime} and AlphaPose~\cite{fang2017rmpe}, and then perform 2D-to-3D lifting to obtain the final estimation results. 
\begin{figure}[t]
\centering
\includegraphics[width=1\columnwidth]{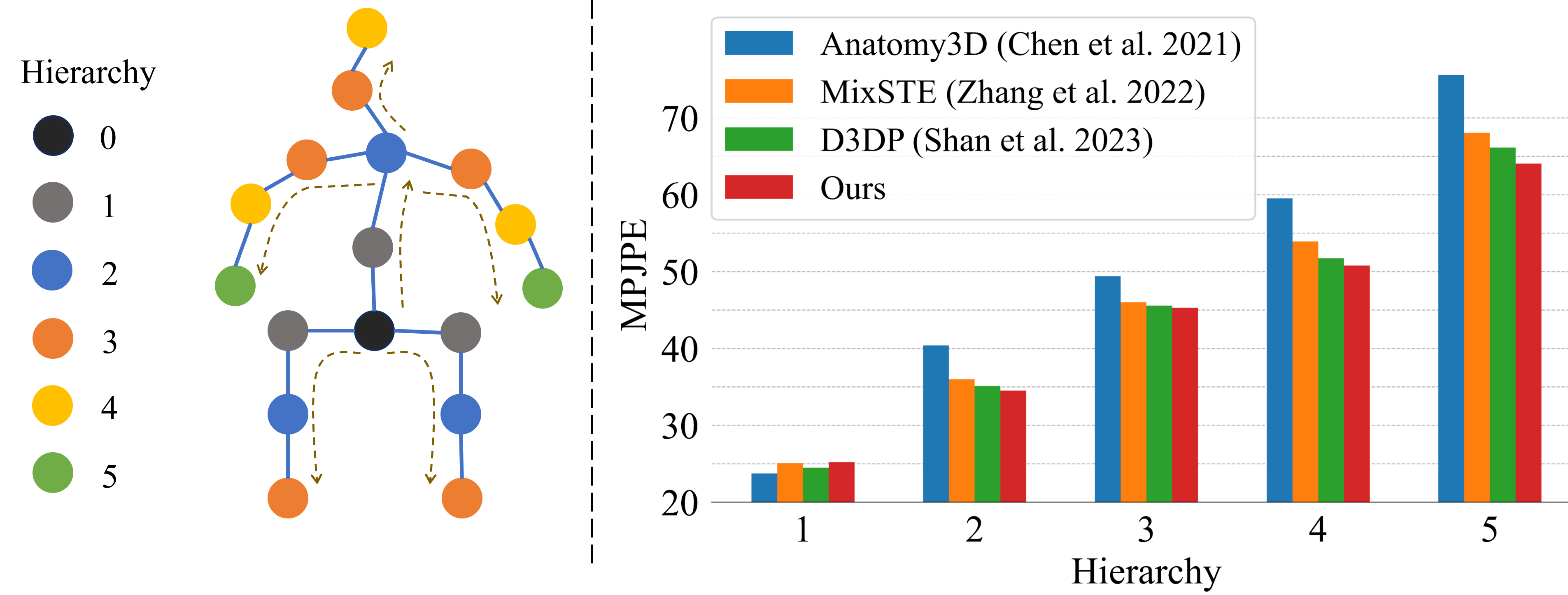} % Reduce the figure size so that it is slightly narrower than the column. Don't use precise values for figure width.This setup will avoid overfull boxes.
\caption{Left: The hierarchy defined in our method and the forward kinematic structure (drawn with brown dashed lines) based on the Human3.6M dataset. Right: The MPJPE of the hierarchy 1-5 joints comparison among Anatomy3D ~\cite{chen2021anatomy}, MixSTE ~\cite{Zhang_2022_CVPR}, D3DP ~\cite{shan2023diffusion} and our method.}
\vspace{-3.85ex}
\label{fig1}
\end{figure}
%The primary challenge in monocular 3D human pose estimation from 2D pose sequence lies in addressing the depth ambiguity. Various potential 3D poses could map the same 2D pose. 
\par Recently, monocular 3D human pose estimation has experienced significant advancements. Many methods have been proposed to alleviate the depth ambiguity. ~\cite{pavllo:videopose3d:2019} considers this issue by exploring temporal information with the convolutional network while the transformer-based methods ~\cite{zheng20213d,Zhang_2022_CVPR} make use of spatial-temporal information to compensate for the information loss in the 2D to 3D mapping process. Learning or introducing human pose prior is another method to mitigate the depth ambiguity. ~\cite{shan2023diffusion,ci2023gfpose,gong2023diffpose} introduce the original pose distribution prior in the training phase, and model 2D-to-3D lifting as a process to denoise from the pose distribution with uncertain noise. Moreover, some disentangle-based methods like ~\cite{xu2020deep,chen2021anatomy,wang2022motion} explicitly predict the bone length and bone direction, subsequently composing 3D joints locations based on the forward kinematics of the human skeleton. Such methods employ explicit pose constraints, integrating symmetry loss, joint angle limits ~\cite{xu2020deep}, and the consistent bone length in the videos ~\cite{chen2021anatomy}.

However, there are two problems existing in these methods: (1) Despite the advantages of disentangle-based techniques in incorporating human pose priors, they come with the drawback of amplified error accumulation, resulting in decreased performance. Meanwhile, diffusion-based 3D HPE methods ~\cite{shan2023diffusion,ci2023gfpose,gong2023diffpose} directly add noise to the original 3D pose which is not conducive to learn the explicit human pose prior. What if we disentangle the diffusion model by adding noise to bone length and direction separately? This disentangle-based model can separately focus on the temporal consistency of bone length and joint angle variations, better enabling the diffusion model to learn human pose prior. (2) Although the transformer-based methods have the ability to explore the spatial-temporal context information, these models generally lack attention to the fine-grained hierarchical information among joints. As shown in the left side of Figure 1, we group joints into six hierarchies based on the kinematic tree depth of the human body. The experiment results in the right side of Figure 1 show a rising hierarchical accumulation error when the hierarchy increases from 1 to 5. 

To solve the problems mentioned above, (1) we introduce the disentangled method in the forward process of diffusion model instead of decomposing the 3D HPE task into bone length and bone direction prediction task, which simplifies learning the human pose prior. (2) For better modeling the hierarchical relation among joints, we propose HSTDenoiser, which contains two modules: the Hierarchical-Related Spatial Transformer (HRST) and the Hierarchical-Related Temporal Transformer (HRTT). In HRST, due to the spatial information of a joint is influenced by its parent joint, we supply the joint's attention with information from its parent joint. Besides, in HRTT, we try to make cross-attention of the joint and the adjacent joints to learn the temporal interrelationships. HRST and HRTT make the joints pay more attention to their hierarchical-related joints, which consequently improves performance on higher-hierarchy joints and contributes to overall better performance.

In conclusion, our contributions can be summarized as follows:
\begin{itemize}
\item We propose the first \textbf{D}isentangled \textbf{D}iffusion-based 3D human \textbf{Pose} Estimation method with \textbf{H}ierarchical Spatial and Temporal Denoiser (\textbf{DDHPose}), which introduces Hierarchical Information in two ways.
\item  We present the Disentangle Strategy for the forward diffusion process based on hierarchical information to better model explicit pose prior. Additionally, we incorporate a disentanglement loss to guide the model's training.
\item The \textbf{HSTDenoiser} is introduced, comprising the Hierarchical-Related Spatial Transformer (\textbf{HRST}) and the Hierarchical-Related Temporal Transformer (\textbf{HRTT}). This denoiser strengthens the relation among the hierarchical joints by enhancing the attention weight of adjacent joints in the reverse diffusion process.
\item Our method outperforms the performance of disentangle-based, non-disentangle based, and probabilistic methods on 3D HPE benchmarks. The qualitative results show that our method has better performance on the higher hierarchy joints.
\end{itemize}

\section{Related Work}
\subsection{3D Human Pose Estimation}
3D HPE can be divided into two categories, one that directly regresses the 3D human pose from raw RGB images and another that first detects the 2D human pose from raw RGB images by using one of the 2D human pose estimation methods like HRNet~\cite{wang2020deep}, CPN~\cite{chen2018cascaded}, OpenPose~\cite{cao2017realtime}, AlphaPose~\cite{fang2017rmpe} and then make a 2D-to-3D lifting to get the final estimation results. ~\cite{tekin2016direct,pavlakos2017coarse,sun2018integral} directly use convolutional neural network to regress 3D pose from a feature volume. Based on the accuracy improvement of 2D human pose estimation, ~\cite{pavllo:videopose3d:2019} uses a fully convolutional model based on dilated temporal convolutions to estimate 3D poses and achieves better results. ~\cite{zheng20213d, Zhang_2022_CVPR, Zhao_2023_CVPR} demonstrate that 3D poses in the video can be effectively estimated with spatial-temporal transformer architecture. Due to the superior performance of two-stage methods, we also employ a two-stage approach for 3D human pose estimation in this paper. While these models are capable of exploring spatial-temporal context information, they always fail to incorporate fine-grained hierarchical information. This leads to a higher hierarchical accumulation error from hierarchy 1 to hierarchy 5 in the right portion of Figure 1. Therefore, we apply HRST and HRTT in our method, providing more hierarchical features for better modeling.

\subsection{Diffusion Model}
The diffusion model belongs to a class of generative models, which has outstanding performance in image generation~\cite{batzolis2021conditional, nichol2021glide, ho2022cascaded}, image super-resolution~\cite{saharia2022image}, semantic segmentation~\cite{baranchuk2021label}, multi-modal tasks~\cite{fan2023frido} and so on.  The diffusion model is first introduced by ~\cite{sohl2015deep}, which defines two stages which are the forward process and the reverse process. The forward process refers to the gradual addition of Gaussian noise to the data until it becomes random noise, while the reverse process is the denoising of noisy data to obtain the true samples. The following works DDPM~\cite{ho2020denoising} and DDIM~\cite{song2020denoising} simplify and accelerate previous diffusion models which make a solid foundation in this area.   
\begin{figure*}[t]
\centering
\includegraphics[scale=0.5]{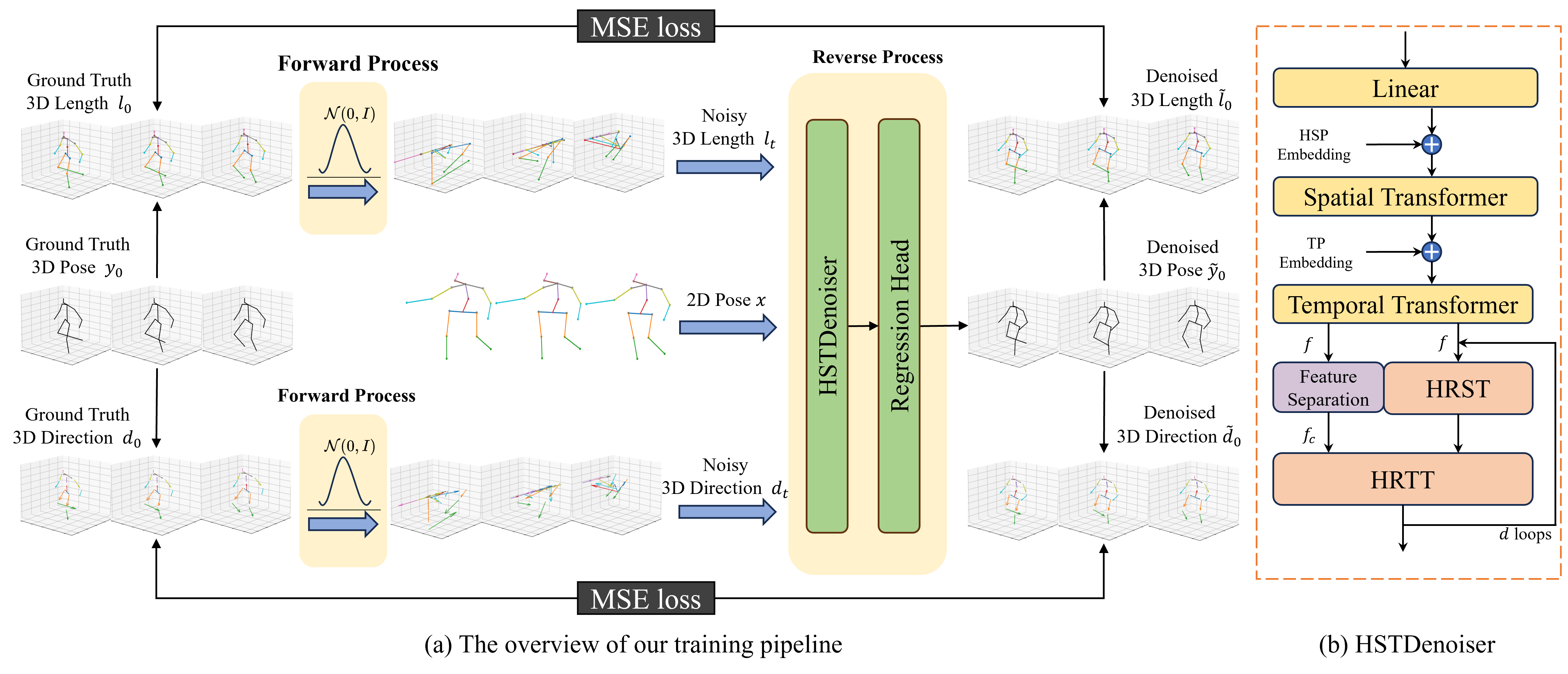} % Reduce the figure size so that it is slightly narrower than the column. Don't use precise values for figure width.This setup will avoid overfull boxes.
\caption{(a): The overview of DDHPose's training pipeline. (b): The architecture of our HSTDenoiser, which contains HRST and HRTT. $HSP$ embedding and $TP$ embedding are used in the spatial-temporal transformer to better modeling the hierarchical relation of spatial position information and temporal position information. $f$ is the feature extracted by the spatial-temporal transformer and $f_{c}$ is the child joint feature separated from $f$. The input consists of $N$ frames for both 2D pose and 3D pose. For better clarity, only three frames of input are illustrated here as an example.}
\label{fig2}
\end{figure*}
\par Recent explorations ~\cite{choi2022diffupose,holmquist2022diffpose,ci2023gfpose,shan2023diffusion} try to apply the diffusion model to 3D human pose estimation. Note that ~\cite{gong2023diffpose} also uses a diffusion model for 3D HPE, but they additionally introduce the heatmap distribution of 2D pose, and the depth distribution to initialize 3D pose distribution, making a GMM-based forward diffusion process, so that they have a better performance than the other diffusion-based 3D HPE model. However, these approaches directly add $t$-step noise in the forward process to the original 3D pose, which is not conducive to learning the explicit human pose prior. Additionally, ~\cite{xu2020deep,chen2021anatomy,wang2022motion} have a higher accumulation of errors that disentangle the 3D joint location to the prediction of bone length and bone direction. We introduce the disentanglement strategy in the forward process of the diffusion model, integrating the explicit human body prior to the diffusion model, and proposing the first disentangle-based diffusion model for 3D HPE. As a result, we achieve outstanding results on 3D HPE benchmarks.

\section{Method}
The overview of our proposed \textbf{DDHPose} is in Figure 2(a). In our framework, we decompose the 3D joint location into the bone length and bone direction, adding noise in the forward process. After the forward process, the noisy bone length, noisy bone direction, and 2D pose are fed to \textbf{HSTDenoiser}, which contains \textbf{HRST} and \textbf{HRTT} to reverse the 3D pose from the noisy input. Further details will be introduced in the following section.

% \subsection{DDHPose}
\subsection{3D POSE Disentanglement Strategy}
% Given the 2D pose sequence condition $x \in\ \mathbb{R}^{N\times J\times2}$ from the 2D pose estimator, our task is to derive the 3D pose sequence $y_{0} \in\ \mathbb{R}^{N\times J\times3}$ from the noisy 3D pose sequence $y_{t} \in\ \mathbb{R}^{N\times J\times3}$, to which a t-step Gaussian noise has been applied. Here, $N$ represents the frame length of the sequence and $J$ denotes the number of human joints. 
We first introduce the motivation of why we use the disentanglement strategy in our paper. The original non-disentangle diffusion-based methods directly take the 3D joint sequence as input without any skeleton structural prior. Modeling the dependencies among each joint pair tends to be challenging due to their complex and dense relation which makes the optimization task more difficult. But in our approach, Our disentangle-based method first decomposes ground truth 3D pose $y_{0} \in\ \mathbb{R}^{N\times J\times3}$ to bone length $l_{0} \in\ \mathbb{R}^{N\times (J-1)\times1}$ and bone direction $d_{0} \in\ \mathbb{R}^{N\times (J-1)\times3}$, where $N$ is the frame length of the input sequences, $J$ is the number of joints. This operation divides the dense and high-dimensional problem into multiple sparse and low-dimensional sub-problems, making the gradient-based optimization easier. Besides, The disentangling representation with bone length and direction makes it easier to add structural constraints, such as temporal consistency in bone length. The addition of bone length loss as a constraint enhances output certainty and shows effectiveness in the experiment.
\par For the $i$-th bone, ground truth length $l_{0}^{i}$ and direction $d_{0}^{i}$ can be defined as:
\begin{equation}
l_{0}^{i} = \left \| y_{0}^{c_{i}} - y_{0}^{p_{i}} \right \|_{2},\quad d_{0}^{i} = \frac{y_{0}^{c_{i}} - y_{0}^{p_{i}}}{\left \| y_{0}^{c_{i}} - y_{0}^{p_{i}} \right \|_{2}}
\end{equation}
where $c_{i}$ and $p_{i}$ are the child joint and parent joint, which are in the upstream and downstream of the $i$-th bone according to the forward kinematic structure defined in the left portion of Figure 1.

\par The disentangled bone length and bone direction are both processed through the forward and reverse processes. 
% The forward process iteratively adds Gaussian noise to the data and gradually converges the data towards a Gaussian distribution. While the reverse process, is designed to denoise the contaminated data by a neural denoiser.
\subsection{The Forward Process}
The Forward Process is an approximate posterior that follows the Markov chain that gradually adds Gaussian noise $\mathcal{N}(0,I)$ to the original data $x_{0}$. Followed by DDPM~\cite{ho2020denoising}, the forward process can be defined as:\\
\begin{equation}
q(x_{t}|x_{0}):=\mathcal{N}(x_{t};\sqrt{\bar{\alpha}_{t}}x_{0},(1-\bar{\alpha}_{t})I) \label{1} \\
\end{equation}
where $\bar{\alpha}_{t}:= {\textstyle \prod_{s=0}^{t}}\alpha_{s}$ and $ \alpha_{s}:=1-\beta_{s}$, $\beta_{s}$ is a noise schedule and we adopt the cosine-schedule proposed by ~\cite{song2020improved} which always increases as the sampling step $t$ increases.

\par During the training stage in Figure 2(a), when we get the disentangled bone length $l_{0}$ and bone direction $d_{0}$, we can do the forward process separately in Eq (2) to get the noisy bone length $l_{t}$ and bone direction $d_{t}$ by adding $t$-step Gaussian noise as:
\begin{equation}
l_{t}=\sqrt{\bar{\alpha}_{t}}l_{0} + \sqrt{1-\bar{\alpha}_{t}}\epsilon,\quad d_{t}=\sqrt{\bar{\alpha}_{t}}d_{0} + \sqrt{1-\bar{\alpha}_{t}}\epsilon
\end{equation}
where $\epsilon$ is the random Gaussian sampled at the $t$-step. 
\subsection{The Reverse Process} 
In the training stage, under the condition of a 2D pose sequence $x \in\ \mathbb{R}^{N\times J\times2}$, the contaminated bone length $l_{t}$ and direction $d_{t}$ from the forward process are concatenated. This combined information is then processed through the HSTDenoiser and a regression head, resulting in the denoised 3D joints locations $\tilde{y}_{0}$. Then using our disentanglement strategy to decompose bone length $\tilde{l}_{0}$ and bone direction $\tilde{d}_{0}$ for disentanglement supervision during training. 

\par At the inference stage, inspired by the method in D3DP~\cite{shan2023diffusion}, we simultaneously sample $H$ hypotheses from the Gaussian distribution as the initial noisy bone length and direction. They are then denoised through the trained denoiser, resulting in the denoised bone length $\tilde{l}_{0:H,0}$ and bone direction $\tilde{d}_{0:H,0}$. Then $\tilde{l}_{0:H,0}$ and $\tilde{l}_{0:H,0}$ are employed to generate noisy samples $\tilde{l}_{0:H,t'}$ and $\tilde{d}_{0:H,t'}$ in the next iteration at step $t'$ via DDIM~\cite{song2020denoising}:
\begin{equation}
\begin{split}
\tilde{l}_{0:H,t'}&=\sqrt{\bar\alpha_{t'}}\tilde{l}_{0:H,0}+\sqrt{1-\bar\alpha_{t'}-\sigma_{t}^{2}}\cdot\epsilon_{tl}+\sigma_{t}\epsilon\\
\tilde{d}_{0:H,t'}&=\sqrt{\bar\alpha_{t'}}\tilde{d}_{0:H,0}+\sqrt{1-\bar\alpha_{t'}-\sigma_{t}^{2}}\cdot\epsilon_{td}+\sigma_{t}\epsilon
\end{split}
\end{equation}
where $\epsilon_{tl}=\frac{\tilde{l}_{0:H,t}-\sqrt{\bar\alpha_{t}}\tilde{l}_{0:H,0}}{\sqrt{1-\bar{\alpha}_{t}}}$, $\epsilon_{td}=\frac{\tilde{d}_{0:H,t}-\sqrt{\bar\alpha_{t}}\tilde{d}_{0:H,0}}{\sqrt{1-\bar{\alpha}_{t}}}$ are the noise at step $t$ and $\sigma_{t}=\sqrt{(1-\bar\alpha_{t^{'}})/(1-\bar\alpha_{t})}\cdot\sqrt{1-\bar\alpha_{t}/\bar\alpha_{t'}}$ controls the stochastic of the diffusion process. We can control the hypothesis number $H$ and iteration times $W$ in the whole process. Appropriately increasing $H$ and $W$ can optimize the final prediction of the bone length and bone direction and improve the performance of MPJPE and P-MPJPE in our experiments.

\subsubsection{Hierarchical Spatial and Temporal Denoiser}
Both in the training or inference phase, noisy bone length and bone direction are fed into our HSTDenoiser to reconstruct the original data. HSTDenoiser, which consists of HRST and HRTT, is used to explore the hierarchical information, specifically the relation among the joint, the parent joint, and the child joint. The main architecture is shown in Figure 2(b). We utilize a linear layer to enhance the input feature and use the spatial-temporal transformer block in MixSTE~\cite{Zhang_2022_CVPR} to extract joint features. We also introduce Hierarchical spatial position embedding $HSP$ for better spatial position modeling and temporal embedding $TP$ for better temporal position modeling. $HSP$ embedding not only contains the spatial position information of each joint but also contains the joint hierarchy information. Inspired by ~\cite{li2023pose}, we split the joints into six hierarchies according to the joint's depth of the human body tree-like structure to build hierarchical embedding, which is shown in the left portion of Figure 1. It means the joints in the same hierarchy share the same embedding. Based on hierarchical embedding, the hierarchical-related information can be well learned by our model. After one layer of spatio-temporal transformer modeling, we utilize the HRST and HRTT, which we introduce in the subsequent section, to model the spatio-temporal correlations of joints through $d$ loops alternately.

\begin{figure}[t]
\centering
\includegraphics[width=1\columnwidth]{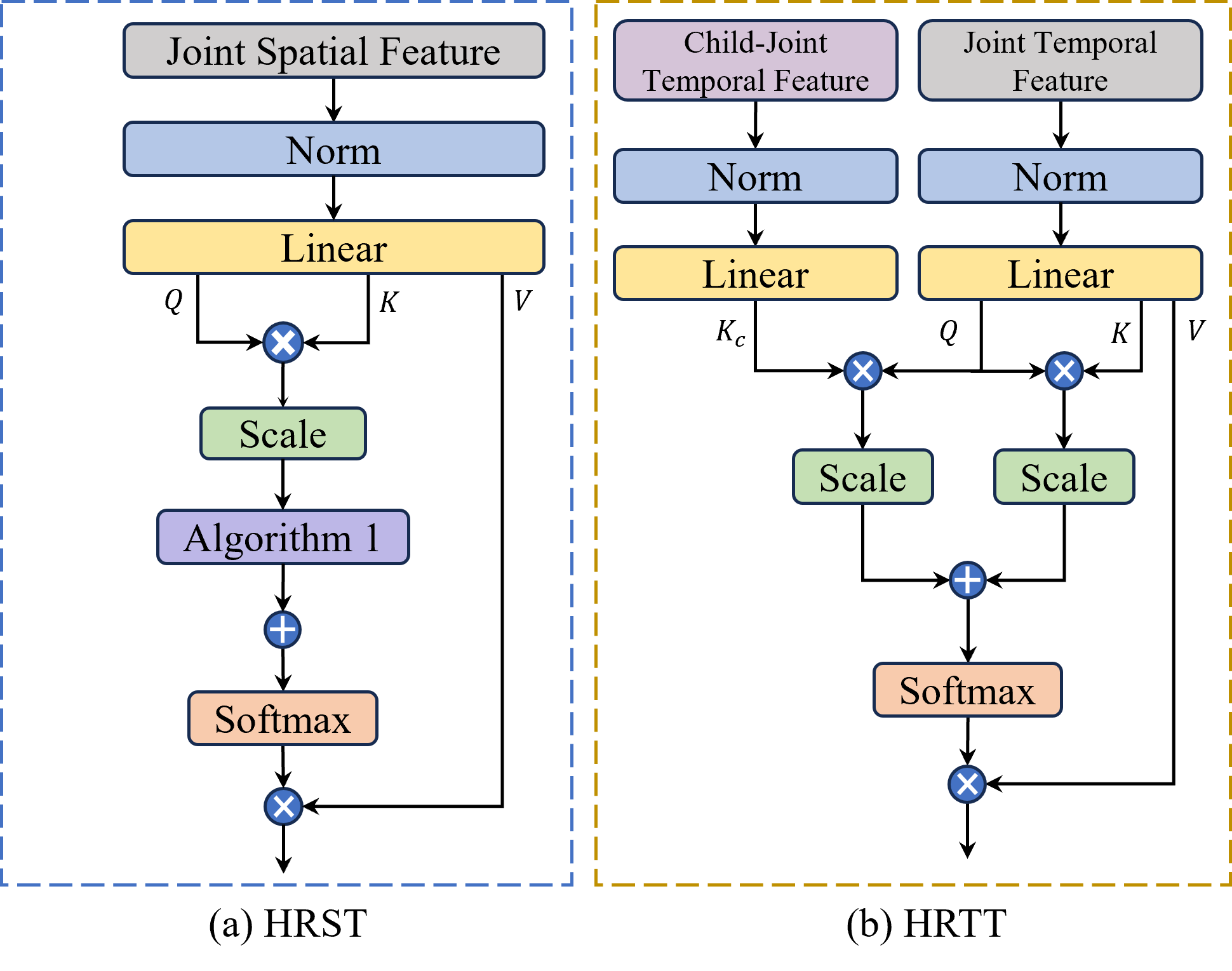} % Reduce the figure size so that it is slightly narrower than the column. Don't use precise values for figure width.This setup will avoid overfull boxes.
\caption{The main components of our HSTDenoiser. (a): Hierarchical-Related Spatial Transformer(HRST). (b): Hierarchical-Related Temporal Transformer(HRTT).}
\label{fig2}
\end{figure}

\subsubsection{Transformer}
The Transformer model we used in our approach is followed by ~\cite{Vaswani_Shazeer_Parmar_Uszkoreit_Jones_Gomez_Kaiser_Polosukhin_2017}, the basic idea of query, key, and value is that the query is used to match with the key, and then according to the degree of matching, to selectively focus on the value. This design allows the output to selectively pay attention to the value based on the query. The mechanism of attention can be formulated by:\\
\begin{equation}
Attention = Softmax(A)V,\quad A = \frac{QK^{T}}{\sqrt{d_{m}}}
\end{equation}
where ${Q,K,V} \in \mathbb{R}^{Z\times d_{m}}$ are generated by the input feature, $Z$ is the number of tokens and $d_{m}$ is the dimension of feature. $A \in \mathbb{R}^{Z\times Z}$ denotes the attention weight matrix.
\par The input of our transformer module is the noisy bone length and direction generated by the forward diffusion process. For better denoising, the 2D pose sequence is added as the condition and concatenated with the 3D noisy data as the whole input. 

\begin{algorithm}[h]
\caption{Hierarchical-Related Spatial Transformer}
\label{alg:algorithm}
\textbf{Input}: $Q$, $K$, $V$ generated by Joint feature $f\in \mathbb{R}^{N\times J\times C}$ \\
\textbf{Parameter}: Hierarchical Related joints triplets $\left \{ J_{p}, J, J_{c} \right \}$\\
\textbf{Output}:\hbox{Hierarchical-Related spatial attention map}
\begin{algorithmic}[1] %[1] enables line numbers
\STATE $A = \frac{QK^{T}}{\sqrt{d_{m}}}$
\FOR{$j_{p}$, $j$, $j_{c}$ in $J_{p}$, $J$, $J_{c}$} 
\STATE $A$[$j$][$j_{c}$] += $A$[$j_{p}$][$j$]
\STATE $A$[$j_{c}$][$j$] += $A$[$j_{p}$][$j$]
\STATE $A$[$j$][$j_{c}$]/2.0 , $A$[$j_{c}$][$j$]/2.0
\ENDFOR
\STATE \textbf{return} $A$
\end{algorithmic}
\end{algorithm}

\subsubsection{HRST}
In HRST, we enhance the modeling of joint spatial information with its parent joint feature. Based on the forward kinematics structure, we define all the hierarchical-related joints triplets in the human body as $\left \{ J_{p}, J, J_{c} \right \}$, where $J_{p}$, $J$, $J_{c}$ are the set of $j_{p}$, $j$, $j_{c}$. In each of the hierarchical-related joints triplets $\left \{ j_{p}, j, j_{c} \right \}$, $j_{p}$ is the parent joint of joint $j$ and $j_{c}$ is the child joint of joint $j$ according to the forward kinematic structure. Because our method decomposes the location of joints into bone length and direction, we believe that the position of a joint is influenced by the position of its parent joint, combined with the bone length and direction. The attention of the parent joint significantly affects the attention of the joint. Therefore, in HRST, we augment the parent joint's influence on each joint at the spatial level, specifically as illustrated in Algorithm 1. After Algorithm 1, we derive the weight matrix $A$ and then compute the attention utilizing Eq (5). 

\subsubsection{HRTT}
We propose HRTT to further introduce the interrelationship between each joint and its hierarchical adjacent joints in the temporal dimension. In the process of exploring joint temporal information, we believe that due to the tree-like structure of the human skeleton, there exists a strong temporal correlation among a joint, its parent joint, and its child joints. Because we have enhanced the relation between a joint and its parent joint, the temporal relation between a joint and its child joints is more considered in HRTT. 
\par Specifically, we primarily adopt a cross-attention mechanism to capture the relationship between the current joint and its child joints. According to the kinematic chain structure of human joints, we compute the average of its features with its child joints as the separated child joint feature $f_{c}$. We use a residual architecture to build the attention weight matrix. For detail, we use joint temporal feature $f$ after HRST to generate $Q$, $K$, $V$, formulating the self-attention weight matrix $A_{s} = \frac{QK^{T}}{\sqrt{d_{m}}}$. Concurrently, in the residual branch, we use the separated child joint feature $f_{c}$ to generate $K_{c}$, making cross-attention weight matrix with $Q$, which can be defined as $A_{c} = \frac{QK_{c}^{T}}{\sqrt{d_{m}}}$. Therefore, $A_{c}$ contains the cross-attention between a joint and its child joints in the temporal dimension. The shape of the temporal attention map $A_{s}$ and $A_{c}$ is $N \times N$, where $N$ is the frame length of the input sequences. $A_{s}$ and $A_{c}$ together dictate the temporal attention focused on $V$ as shown in Eq (6). Since $A_{s}$ contains the enhanced weight of both the joint and its parent joint feature, $A_{c}$ in the residual branch augments the relation between a joint and its child joints. Hence, HRTT effectively captures the relationships between the hierarchical adjacent joints.
\begin{equation}
\begin{split}
    Attention_{HRTT} &= Softmax(A_{s} + A_{c})V\\
\end{split}
\end{equation}

\begingroup
\setlength{\tabcolsep}{1.2pt} % Default value: 6pt
\begin{table*}[t]
\footnotesize
\begin{tabular}{lcccccccccccccccc}
\hline
\multicolumn{17}{c}{\textbf{Deterministic methods: Disentangle-based model}}                                                                                                                                      \\ \hline
\multicolumn{1}{l|}{\textbf{Protocol \#1: MPJPE}}         & Dir. & Disc. & Eat  & Greet & Phone & Photo & Pose & Pur. & Sit  & SitD. & Smoke & Wait & WalkD. & Walk & \multicolumn{1}{c|}{WalkT.} & \textbf{Avg.} \\ \hline
\multicolumn{1}{l|}{DKA~\cite{xu2020deep}($N$=9)} & 37.4          & 43.5          & 42.7          & 42.7          & 46.6          & 59.7          & 41.3          & 45.1          & 52.7          & 60.2          & 45.8          & 43.1          & 47.7          & 33.7          & \multicolumn{1}{c|}{37.1}          & 45.6          \\
\multicolumn{1}{l|}{Anatomy3D~\cite{chen2021anatomy}($N$=243)}     & 41.4          & 43.5          & 40.1          & 42.9          & 46.6          & 51.9          & 41.7          & 42.3          & 53.9          & 60.2          & 45.4          & 41.7          & 46.0          & 31.5          & \multicolumn{1}{c|}{32.7}          & 44.1          \\
\multicolumn{1}{l|}{Virtual Bones~\cite{wang2022motion}($N$=243)}      & 42.4          & 43.5          & 41.0          & 43.5          & 46.7          & 54.6          & 42.5          & 42.1          & 54.9          & 60.5          & 45.7          & 42.1          & 46.5          & 31.7          & \multicolumn{1}{c|}{33.7}          & 44.8          \\ \hline
\multicolumn{1}{l|}{Ours ($N$=243, $H$=1, $W$=1)}                & \textbf{37.3}        & \textbf{40.0}        & \textbf{35.2}        & \textbf{37.7}        & \textbf{41.1}        & \textbf{46.7}        & \textbf{38.4}        & \textbf{38.4}        & \textbf{52.2}        & \textbf{53.3}        & \textbf{41.4}        & \textbf{38.9}        & \textbf{38.8}        & \textbf{27.6}        & \multicolumn{1}{c|}{\textbf{27.7}} & \textbf{39.7}        \\ \hline
                                                          &      &       &      &       &       &       &      &      &      &       &       &      &        &      &                             &               \\ \hline

\multicolumn{17}{c}{\textbf{Deterministic methods: Non-Disentangle based model}}                                                                                                                                                                                                                                                                                                                                                                                                                                                                                            \\ \hline
\multicolumn{1}{l|}{\textbf{Protocol \#1: MPJPE}}         & Dir.                        & Disc.                       & Eat                         & Greet                       & Phone                       & Photo                       & Pose                        & Pur.                        & Sit                         & SitD.                       & Smoke                       & Wait                        & WalkD.                      & Walk                        & \multicolumn{1}{c|}{WalkT.}                      & \textbf{Avg.}               \\ \hline
\multicolumn{1}{l|}{VideoPose3D~\cite{pavllo:videopose3d:2019}($N$=243)}        & 45.2          & 46.7          & 43.3          & 45.6          & 48.1          & 55.1          & 44.6          & 44.3          & 57.3          & 65.8          & 47.1          & 44.0          & 49.0          & 32.8          & \multicolumn{1}{c|}{33.9}          & 46.8          \\
\multicolumn{1}{l|}{PoseFormer~\cite{zheng20213d}($N$=81)}      & 41.5          & 44.8          & 39.8          & 42.5          & 46.5          & 51.6          & 42.1          & 42.0          & 53.3          & 60.7          & 45.5          & 43.3          & 46.1          & 31.8          & \multicolumn{1}{c|}{32.2}          & 44.3          \\
\multicolumn{1}{l|}{P-STMO~\cite{shan2022p}($N$=243)}      & 38.9          & 42.7          & 40.4          & 41.1          & 45.6          & 49.7          & 40.9          & 39.9          & 55.5          & 59.4          & 44.9          & 42.2          & 42.7          & 29.4          & \multicolumn{1}{c|}{29.4}          & 42.8          \\
\multicolumn{1}{l|}{MixSTE~\cite{Zhang_2022_CVPR}($N$=243)}    & 37.6          & 40.9          & 37.3          & 39.7          & 42.3          & 49.9          & 40.1          & 39.8          & 51.7          & 55.0          & 42.1          & 39.8          & 41.0          & 27.9          & \multicolumn{1}{c|}{27.9}          & 40.9          \\
\multicolumn{1}{l|}{PoseFormerV2~\cite{Zhao_2023_CVPR}($N$=243)}         & 41.3          & 45.5          & 41.5          & 44.0          & 46.7          & 53.8          & 42.6          & 42.6          & 55.2          & 64.6          & 45.7          & 42.9          & 45.8          & 32.3          & \multicolumn{1}{c|}{32.9}          & 45.2          \\
\multicolumn{1}{l|}{STCFormer~\cite{tang20233d}($N$=243)}       & 38.4          & 41.2          & 36.8          & 38.0          & 42.7          & 50.5          & 38.7          & \textbf{38.2} & 52.5          & 56.8          & 41.8          & \textbf{38.4} & 40.2          & \textbf{26.2} & \multicolumn{1}{c|}{\textbf{27.7}} & 40.5          \\ \hline
\multicolumn{1}{l|}{Ours ($N$=243, $H$=1, $W$=1)}                & \textbf{37.3} & \textbf{40.0} & \textbf{35.2} & \textbf{37.7} & \textbf{41.1} & \textbf{46.7} & \textbf{38.4} & 38.4          & \textbf{52.2} & \textbf{53.3} & \textbf{41.4} & 38.9          & \textbf{38.8} & 27.6          & \multicolumn{1}{c|}{\textbf{27.7}} & \textbf{39.7} \\ \hline
                                                          & \multicolumn{1}{l}{}        & \multicolumn{1}{l}{}        & \multicolumn{1}{l}{}        & \multicolumn{1}{l}{}        & \multicolumn{1}{l}{}        & \multicolumn{1}{l}{}        & \multicolumn{1}{l}{}        & \multicolumn{1}{l}{}        & \multicolumn{1}{l}{}        & \multicolumn{1}{l}{}        & \multicolumn{1}{l}{}        & \multicolumn{1}{l}{}        & \multicolumn{1}{l}{}        & \multicolumn{1}{l}{}        & \multicolumn{1}{l}{}                             & \multicolumn{1}{l}{}  \\ \hline                                                                                                                                                                                                                                                                                                                                                                                                                                                                                              
\multicolumn{17}{c}{\textbf{Probabilistic methods}}                                                                                                                                                                                                                                                                                                                                                                                                                                                                                                                     \\ \hline
\multicolumn{1}{l|}{\textbf{Protocol \#1: MPJPE}} & Dir.                        & Disc.                       & Eat                         & Greet                       & Phone                       & Photo                       & Pose                        & Pur.                        & Sit                         & SitD.                       & Smoke                       & Wait                        & WalkD.                      & Walk                        & \multicolumn{1}{c|}{WalkT.}                      & \textbf{Avg.}               \\ \hline
\multicolumn{1}{l|}{MHFormer~\cite{li2022mhformer}($N$ =351, $H$=3)} & 39.2          & 43.1          & 40.1          & 40.9          & 44.9          & 51.2          & 40.6          & 41.3          & 53.5          & 60.3          & 43.7          & 41.1          & 43.8          & 29.8          & \multicolumn{1}{c|}{30.6}          & 43.0          \\
\multicolumn{1}{l|}{GFPose~\cite{ci2023gfpose}($H$ = 10)}              & 39.9          & 44.6          & 40.2          & 41.3          & 46.7          & 53.6          & 41.9          & 40.4          & 52.1          & 67.1          & 45.7          & 42.9          & 46.1          & 36.5          & \multicolumn{1}{c|}{38.0}          & 45.1          \\
\multicolumn{1}{l|}{D3DP~\cite{shan2023diffusion}($N$=243,$*$)}      & 37.3          & \textbf{39.5} & 35.6          & 37.8          & 41.3          & 48.2          & 39.1          & \textbf{37.6} & \textbf{49.9} & 52.8          & 41.2          & 39.2          & 39.4          & 27.2          & \multicolumn{1}{c|}{27.1}          & 39.5          \\ \hline
\multicolumn{1}{l|}{Ours ($N$=243, $H$=5, $W$=1)}        & 37.2          & 39.9          & 35.1          & \textbf{37.6} & 41.0          & 46.5          & 38.3          & 38.3          & 52.1          & 53.1          & 41.3          & 38.8          & 38.7          & 27.5          & \multicolumn{1}{c|}{27.6}          & 39.5          \\
\multicolumn{1}{l|}{Ours ($N$=243, $H$=20, $W$=10)}      & \textbf{36.4} & \textbf{39.5} & \textbf{34.9} & \textbf{37.6} & \textbf{40.1} & \textbf{45.9} & \textbf{37.8} & 37.8          & 51.5          & \textbf{52.2} & \textbf{40.8} & \textbf{38.3} & \textbf{38.3} & \textbf{27.0} & \multicolumn{1}{c|}{\textbf{27.0}} & \textbf{39.0} \\ \hline
\end{tabular}
\caption{Results on Human3.6M in millimeters under MPJPE. $N$, $H$, $W$: the number of input frames, hypotheses, and iterations used in the inference stage. In this table, we compare with the deterministic and probabilistic methods. The best results are highlighted in bold. ($*$)-For clarity, $H$=20, $W$=10 is omitted. }
\end{table*}
\endgroup

\subsection{Loss Function} 
\subsubsection{3D Disentanglement Loss}
3D disentanglement loss is utilized to aid the model in learning the explicit priors during the forward diffusion process. Given the 3D ground truth pose sequence $y_{0}$ and the predicted 3D pose sequence $\tilde{y}_{0}$ , we decompose $y_{0}$ to bone length $l_{0}$ and bone direction $d_{0}$. Similarly, we can obtain the disentangled bone length prediction $\tilde{l}_{0}$ and bone direction prediction $\tilde{d}_{0}$. And for the $i$-th bone, length $l_{0}^{i}$, $\tilde{l}_{0}^{i}$ and direction $d_{0}^{i}$, $\tilde{d}_{0}^{i}$ are defined as:

\begin{equation}
\begin{aligned}
l_{0}^{i} &= \left \| y_{0}^{c_{i}} - y_{0}^{p_{i}} \right \|_{2}, &\tilde{l}_{0}^{i} &= \left \| \tilde{y}_{0}^{c_{i}} - \tilde{y}_{0}^{p_{i}} \right \|_{2}\\
d_{0}^{i} &= \frac{y_{0}^{c_{i}} - y_{0}^{p_{i}}}{\left \| y_{0}^{c_{i}} - y_{0}^{p_{i}} \right \|_{2}}, &\tilde{d}_{0}^{i} &= \frac{\tilde{y}_{0}^{c_{i}} - \tilde{y}_{0}^{p_{i}}}{\left \| \tilde{y}_{0}^{c_{i}} - \tilde{y}_{0}^{p_{i}} \right \|_{2}}
\end{aligned}
\end{equation}
where $c_{i}$ and $p_{i}$ is the child joint and parent joint of the $i$-th bone. Then the disentanglement loss we use in our training stage can be defined as:\\
\begin{equation}
\begin{gathered}
\begin{aligned}
L_{l} &= \left \| \tilde{l}_{0} - l_{0} \right \|_{2}, & L_{d} &= \left \| \tilde{d}_{0} - d_{0} \right \|_{2}\\
\end{aligned} \\
L_{dis} = L_{l} + L_{d}
\end{gathered}
\end{equation}

\subsubsection{3D Pose Loss}
During the model's training process, we also use 3D pose loss to constrain the denoised 3D pose regressed by the model: 
\begin{equation}
    L_{pos} = \left \| \tilde{y}_{0} - y_{0} \right \|_{2}
\end{equation}

By combining the 3D disentanglement loss and the 3D pose loss, the overall loss we used to supervise our model is given as:\\
\begin{equation}
    L = L_{dis} + L_{pos}
\end{equation}

\section{Experiments}

\subsection{Dataset}
\subsubsection{Human3.6M} ~\cite{h36m_pami} is widely used in 3D HPE task. It contains 3.6 million 3D human poses and corresponding images with 11 professional actors and collected in 17 scenarios. Following the previous work ~\cite{pavllo:videopose3d:2019,zheng20213d,Zhang_2022_CVPR}, we use S1-S9 for training and use S9 and S11 for testing.

\subsubsection{MPI-INF-3DHP} ~\cite{mono-3dhp2017} record 8 actors, composed of 4 males and 4 females, each undertaking 8 different sets of activities. We use eight activities performed by eight actors to train our model, while the test dataset has seven different activities.
\subsection{Metrics}
We use the mean per joint position error (MPJPE) and procrustes mean per joint position error(P-MPJPE) for evaluation. MPJPE measures the Euclidean distance between the ground truth and the predicted 3D positions of each joint while P-MPJPE makes procrustes analysis involves scaling, translating, and rotating the predicted pose to best align it with the ground truth, providing a more fair comparison. Following D3DP~\cite{shan2023diffusion}, we use J-AGG based MPJPE and P-MPJPE to evaluate our results.

\subsection{Comparison with State-of-the-art Methods}
\subsubsection{Results on Human3.6M}
The results of our method on Human3.6M are presented in Table 1. We first compare our method with the SOTA deterministic 3D human pose estimation methods. Based on whether the regression of the 3D pose locations is decomposed into the regression of bone length and bone direction, we divide the methods into disentangle-based methods and non-disentangle based methods. For disentangle-based methods, we can see from the table that our method achieves the best MPJPE of 39.7mm, surpassing Anatomy3D~\cite{chen2021anatomy} by 4.4mm(10.0\%) in MPJPE. For non-disentangle based model, we improve STCFormer~\cite{tang20233d} by 0.8mm(2.0\%) under MPJPE. And then we compare our method with probabilistic methods, our method reaches the SOTA MPJPE of 39.0mm, outerperforms  D3DP ~\cite{shan2023diffusion} by 0.5mm(1.3\%). 
\begingroup
\setlength{\tabcolsep}{2pt}
\begin{table}[t]
\centering
\footnotesize
\begin{tabular}{l|cc}
\hline
\multicolumn{1}{c|}{Method}                  & MPJPE & P-MPJPE \\ \hline
DiffPose~\cite{gong2023diffpose}($N$=243, $\flat $)            & 36.9  & 28.7    \\
DiffPose$\sharp$~\cite{gong2023diffpose}($N$=243, $\flat $)            & 40.1  & 31.1    \\ \hline
Ours ($N$=243, $H$=5, $W$=50) & 39.2  & 31.1    \\ \hline
\end{tabular}
\caption{Comparison with DiffPose~\cite{gong2023diffpose} on Human3.6M. ($\sharp$)- Stand-Diff implemented in DiffPose. ($\flat $)-For clarity, $H$=5, $W$=50 is omitted.}
\end{table}
\endgroup

\begingroup
\setlength{\tabcolsep}{2pt}
\begin{table}[t]
\centering
\footnotesize
\begin{tabular}{l|ccc}
\hline
\multicolumn{1}{c|}{Method}       & PCK$\uparrow$  & AUC$\uparrow$  & MPJPE$\downarrow$ \\ \hline
Anatomy3D~\cite{chen2021anatomy}($N$=243)            & 87.8 & 53.8 & 79.1  \\
PoseFormer~\cite{zheng20213d}($N$=9)            & 88.6 & 56.4 & 77.1  \\
P-STMO~\cite{shan2022p}($N$=81)            & 97.9 & 75.8 & 32.2  \\
MixSTE~\cite{Zhang_2022_CVPR}($N$=243)           & 96.9 & 75.8 & 35.4  \\
D3DP~\cite{shan2023diffusion}($N$=243,$\natural$)        & 97.7 & 77.8 & 30.2  \\ \hline
Ours ($N$=243,$\natural$) & \textbf{98.5} & \textbf{78.1} & \textbf{29.2}  \\ \hline
\end{tabular}
\caption{Results on MPI-INF-3DHP under PCK, AUC, and MPJPE using ground truth 2D pose as inputs. The best results are highlighted in bold. ($\natural $)-For clarity, $H$=1, $W$=1 is omitted.}
\end{table}
\endgroup

As for DiffPose~\cite{gong2023diffpose}, we separately compare with it in Table 2. Note that, the DiffPose additionally introduces the heatmaps derived from an off-the-shelf 2D pose detector and depth distributions to initialize the pose distribution. The probabilistic methods in Table 1 only use the 2D pose sequences. Thus, it might not be fair to directly compare with DiffPose. But according to DiffPose, the implementation of Stand-Diff only uses 2D pose sequences by reversing the 3D pose from a standard Gaussian noise, which achieves a larger MPJPE error than our DDHPose with the same setting (40.1mm vs 39.2mm). The results demonstrate that our method can notably boost performance by 0.9mm through the Disentangle Strategy and the utilization of hierarchical relations.

\subsubsection{Results on MPI-INF-3DHP}
We also evaluate our method on the MPI-INF-3DHP dataset under PCK, AUC, and MPJPE metrics. In Table 3, our approach outperforms the SOTA method by 0.8 in PCK, 0.3 in AUC, and 1.0mm in MPJPE under the single hypothesis condition.

\begingroup
\setlength{\tabcolsep}{1.5pt}
\begin{table}[h]
\begin{tabular}{cc|cc}
\hline
Disentangled Input & Disentangled Output & MPJPE & P-MPJPE \\ \hline
\ding{55}                  & \ding{55}              & 40.23  & 31.56   \\
\ding{55}                  & \ding{51}              & 41.72      & 33.06        \\
\ding{51}             & \ding{55}                   & \textbf{39.65} & \textbf{31.24}   \\
\ding{51}              & \ding{51}                  & 40.48 & 32.08   \\ \hline
\end{tabular}
\caption{The impact of disentanglement strategy. The disentanglement strategy with Disentangled input and without Disentangled output has the best result highlighted in bold.}
\end{table}
\endgroup

\subsection{Ablation Study}
In order to evaluate each design in our method, we conduct ablation experiments on the Human3.6M dataset using 2D pose sequence extracted by CPN. 

\subsubsection{Impact of Disentanglement Strategy}
In this section, we separately compare the effect of the Disentangle Input and Disentangle Output strategy. 
\par For the Disentangle Input Strategy, our method divides the dense and high-dimensional optimization problem into two low-dimensional sub-problems, simplifying the learning of the human pose prior. As shown in the left portion of Figure 4, employing the Disentangle Input strategy results in faster convergence and lower training 3D pose loss compared to not using it in the initial training epoch. This leads to improved quantitative results (39.65mm vs 40.23mm), as highlighted in Table 4.
\par For Disentangle Output, the denoiser in the reverse process directly regresses bone length and direction, generating the 3D pose using $C = C_{p}+l \times d$, where $C$ and $C_{p}$ are joint and parent joint coordinates, and $l$, $d$ represent predicted bone length and direction. This equation indicates that a joint's coordinate depends not only on its own bone properties but also on all parent joints along the bone chain. As illustrated in the right portion of Figure 4, hierarchy 1 exhibits lower errors in the Disentangled Output setting, while higher hierarchical levels accumulate errors more than without using Disentangled Output. Quantitative results in Table 4 show that employing the Disentangle Output strategy increases MPJPE from 40.23mm to 41.72mm.

\begin{figure}[b]
\centering
\includegraphics[width=1\columnwidth]{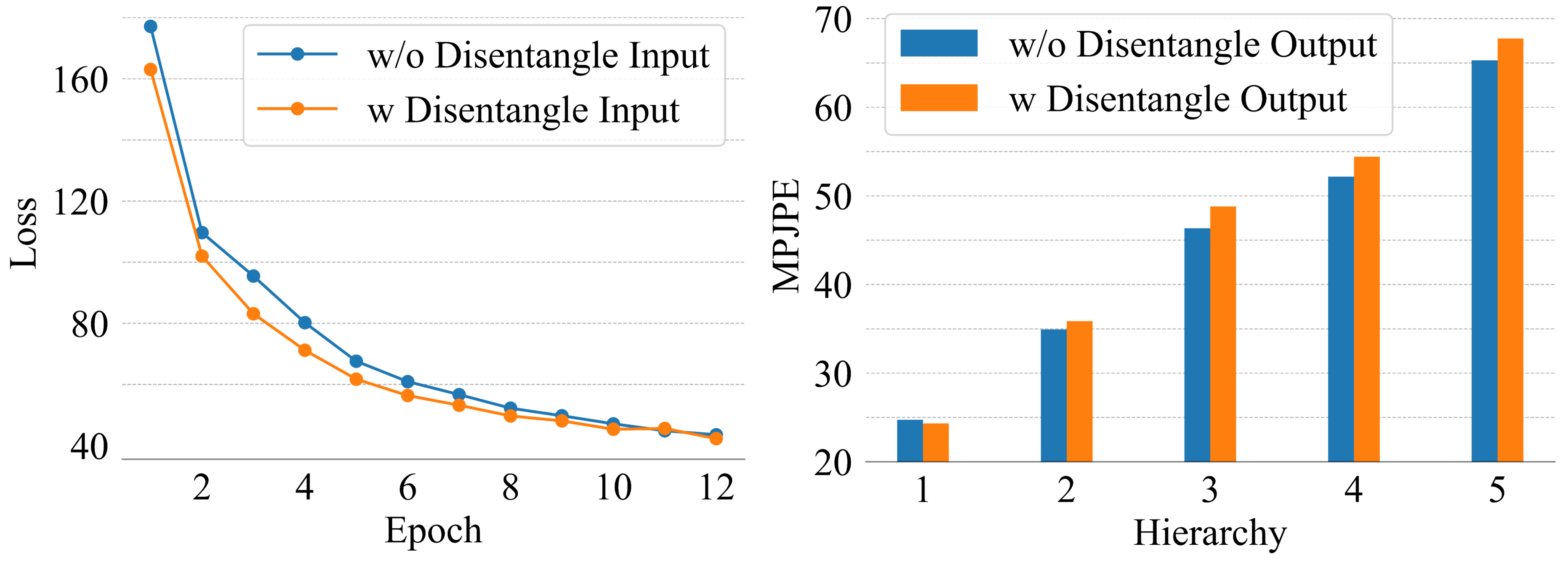} % Reduce the figure size so that it is slightly narrower than the column. Don't use precise values for figure width.This setup will avoid overfull boxes.
\caption{Left: Training Loss Comparison (w/o Disentangle Output). Right: Hierarchical Error Comparison (w/o Disentangle Input).}
\label{fig2}
\end{figure}

\subsubsection{Effect of each module}
As shown in Table 5. We can divide our method into three modules: Hierarchical embedding, HRST, and HRTT. In our experiment, we sequentially add the modules to the baseline, which doesn't use any of the three modules to verify the effectiveness of each module. For simplicity, we set both $H$ and $W$ to 1. 

\begingroup
\setlength{\tabcolsep}{2.5pt}
\begin{table}[t]
\begin{tabular}{cccc|cc}
\hline
baseline & \begin{tabular}[c]{@{}c@{}}Hierarchical \\ embedding\end{tabular} & HRST & HRTT & MPJPE          & P-MPJPE        \\ \hline
\ding{51} &      &      &      & 40.10          & 31.94          \\
\ding{51} & \ding{51}    &      &      & 40.08          & 31.83          \\
\ding{51} & \ding{51}    & \ding{51}    &      & 39.68          & 31.55          \\
\ding{51} & \ding{51}    & \ding{51} & \ding{51}    & \textbf{39.65} & \textbf{31.24} \\ \hline
\end{tabular}
\caption{Effect of each module in our experiments on Human3.6M dataset. The best results are highlighted in bold.}
\end{table}
\endgroup

The result shows that hierarchical embedding provides a slight improvement over the baseline. Adding HRST can boost MPJPE from 40.08mm to 39.68mm and lifts P-MPJPE from 31.83mm to 31.55mm. Further integrating HRTT refines the MPJPE from 39.68mm to 39.65mm and boosts P-MPJPE from 31.55mm to 31.24mm. The results suggest that information from the parent joint influences the regression of the joint itself, which assists the model in learning the joint's spatial information. The shared information between parent and child joints aids the model in inferring the temporal feature of the joint.

\subsubsection{Effect of Loss Function}
We employ the 3D pose loss to constrain the denoised 3D pose regressed by our model and utilize the 3D disentanglement loss to aid the model in learning the explicit human body prior during the forward diffusion process. The contribution of the loss function is in Table 6. The result shows that using 3D disentanglement loss is essential for a better result. With the 3D disentanglement loss, the performance of MPJPE and P-MPJPE can be improved by 0.22mm and 0.70mm.
\begingroup
\setlength{\tabcolsep}{4pt}
\begin{table}[h]
\centering
\begin{tabular}{l|cc}
\hline
\multicolumn{1}{c|}{}               & MPJPE & P-MPJPE \\ \hline
3D pos loss                        & 39.87 & 31.94   \\
3D pos loss + 3D dis loss & \textbf{39.65} & \textbf{31.24}   \\ \hline
\end{tabular}
\caption{Ablation study for loss function proposed in our method. The best results are highlighted in bold.}
\end{table}
\endgroup

\section{Conclusion}
In this paper, we propose DDHPose, a diffusion-based 3D HPE method that introduces hierarchical information in two ways: (1)We propose the Disentangle Strategy for the forward diffusion process, which decomposes the 3D pose into bone length and direction based on the Hierarchical Information. This simplifies learning the human pose prior, reduces optimization dimension, and speeds up gradient descent. (2)We propose HSTDenoiser to strengthen the relation among the hierarchical joints by enhancing the attention weight of adjacent joints for each joint in the reverse diffusion process. Extensive results on Human3.6M and MPI-INF-3DHP reveal that our method surpasses the disentangle-based method, non-disentangle based method, and the probabilistic approaches on 3D HPE benchmarks.

\section{Acknowledgments}
This work is jointly supported by National Natural Science Foundation of China (62276025, 62206022), Beijing Municipal Science \& Technology Commission (Z231100007423015) and Shenzhen Technology Plan Program (KQTD20170331093217368).

\section{Appendix}

\begingroup
\setlength{\tabcolsep}{1.2pt}
\begin{table*}[h]
\footnotesize
\begin{tabular}{lcccccccccccccccc}
\hline
\multicolumn{17}{c}{\textbf{Deterministic methods: Disentangle-based model}}                                                                                                                                                                                                                                                                                                                                                                                                                                                                                                    \\ \hline
\multicolumn{1}{l|}{\textbf{Protocol \#2: P-MPJPE}}       & Dir.                        & Disc.                       & Eat                         & Greet                       & Phone                       & Photo                       & Pose                        & Pur.                        & Sit                         & SitD.                       & Smoke                       & Wait                        & WalkD.                      & Walk                        & \multicolumn{1}{c|}{WalkT.}                      & \textbf{Avg.}               \\ \hline
\multicolumn{1}{l|}{DKA~\cite{xu2020deep}($N$=9)} & 31.0                 & 34.8                 & 34.7                 & 34.4                 & 36.2                 & 43.9                 & 31.6                 & 33.5                 & 42.3                 & 49.0                 & 37.1                 & 33.0                 & 39.1                 & 26.9                 & \multicolumn{1}{c|}{31.9}          & 36.2                 \\
\multicolumn{1}{l|}{Anatomy3D~\cite{chen2021anatomy}($N$=243)}     & 32.6                 & 35.1                 & 32.8                 & 35.4                 & 36.3                 & 40.4                 & 32.4                 & 32.3                 & 42.7                 & 49.0                 & 36.8                 & 32.4                 & 36.0                 & 24.9                 & \multicolumn{1}{c|}{26.5}          & 35.0                 \\
\multicolumn{1}{l|}{Virtual Bones~\cite{wang2022motion}($N$=243)}      & 32.2                 & 34.9                 & 33.0                 & 35.2                 & 35.7                 & 40.7                 & 32.6                 & 32.1                 & 42.8                 & 48.9                 & 36.5                 & 32.5                 & 35.9                 & 25.0                 & \multicolumn{1}{c|}{26.7}          & 34.9                 \\ \hline
\multicolumn{1}{l|}{Ours($N$=243, $H$=1, $W$=1)}                & \textbf{29.5}        & \textbf{31.9}        & \textbf{28.7}        & \textbf{30.1}        & \textbf{31.3}        & \textbf{35.5}        & \textbf{29.7}        & \textbf{29.5}        & \textbf{41.8}        & \textbf{42.9}        & \textbf{33.4}        & \textbf{29.7}        & \textbf{30.4}        & \textbf{21.7}        & \multicolumn{1}{c|}{\textbf{22.6}} & \textbf{31.2}        \\ \hline
                                                          & \multicolumn{1}{l}{}        & \multicolumn{1}{l}{}        & \multicolumn{1}{l}{}        & \multicolumn{1}{l}{}        & \multicolumn{1}{l}{}        & \multicolumn{1}{l}{}        & \multicolumn{1}{l}{}        & \multicolumn{1}{l}{}        & \multicolumn{1}{l}{}        & \multicolumn{1}{l}{}        & \multicolumn{1}{l}{}        & \multicolumn{1}{l}{}        & \multicolumn{1}{l}{}        & \multicolumn{1}{l}{}        & \multicolumn{1}{l}{}                             & \multicolumn{1}{l}{}        \\ \hline
\multicolumn{17}{c}{\textbf{Deterministic methods: Non-Disentangle-based model}}                                                                                                                                                                                                                                                                                                                                                                                                                                                                                               \\ \hline
\multicolumn{1}{l|}{\textbf{Protocol \#2: P-MPJPE}}       & Dir.                        & Disc.                       & Eat                         & Greet                       & Phone                       & Photo                       & Pose                        & Pur.                        & Sit                         & SitD.                       & Smoke                       & Wait                        & WalkD.                      & Walk                        & \multicolumn{1}{c|}{WalkT.}                      & \textbf{Avg.}               \\ \hline
\multicolumn{1}{l|}{VideoPose3D~\cite{pavllo:videopose3d:2019}($N$=243)}        & 34.1                 & 36.1                 & 34.4                 & 37.2                 & 36.4                 & 42.2                 & 34.4                 & 33.6                 & 45.0                 & 52.5                 & 37.4                 & 33.8                 & 37.8                 & 25.6                 & \multicolumn{1}{c|}{27.3}          & 36.5                 \\
\multicolumn{1}{l|}{PoseFormer~\cite{zheng20213d}($N$=81)}      & 34.1                 & 36.1                 & 34.4                 & 37.2                 & 34.4                 & 39.2                 & 32.0                 & 31.8                 & 42.9                 & 46.9                 & 35.5                 & 32.0                 & 34.4                 & 23.6                 & \multicolumn{1}{c|}{25.2}          & 33.9                 \\
\multicolumn{1}{l|}{P-STMO~\cite{shan2022p}($N$ =243)}      & 31.3                 & 35.2                 & 32.9                 & 33.9                 & 35.4                 & 39.3                 & 32.5                 & 31.5                 & 44.6                 & 48.2                 & 36.3                 & 32.9                 & 34.4                 & 23.8                 & \multicolumn{1}{c|}{23.9}          & 34.4                 \\
\multicolumn{1}{l|}{MixSTE~\cite{Zhang_2022_CVPR}($N$ =243)}    & 30.8                 & 33.1                 & 30.3                 & 31.8                 & 33.1                 & 39.1                 & 31.1                 & 30.5                 & 42.5                 & 44.5                 & 34.0                 & 30.8                 & 32.7                 & 22.1                 & \multicolumn{1}{c|}{22.9}          & 32.6                 \\
\multicolumn{1}{l|}{PoseFormerV2~\cite{Zhao_2023_CVPR}($N$ =243)}        & 32.3                 & 35.9                 & 33.8                 & 35.8                 & 36.0                 & 41.1                 & 33.2                 & 32.7                 & 44.3                 & 51.9                 & 37.4                 & 32.8                 & 35.6                 & 25.2                 & \multicolumn{1}{c|}{26.6}          & 35.6                 \\
\multicolumn{1}{l|}{STCFormer~\cite{tang20233d}($N$=243)}       & \textbf{29.3}        & 33.0                 & 30.7                 & 30.6                 & 32.7                 & 38.2                 & \textbf{29.7}        & \textbf{28.8}        & 42.2                 & 45.0                 & \textbf{33.3}        & \textbf{29.4}        & 31.5                 & \textbf{20.9}        & \multicolumn{1}{c|}{\textbf{22.3}} & 31.8                 \\ \hline
\multicolumn{1}{l|}{Ours($N$=243, $H$=1, $W$=1)}                & 29.5                 & \textbf{31.9}        & \textbf{28.7}        & \textbf{30.1}        & \textbf{31.3}        & \textbf{35.5}        & \textbf{29.7}        & 29.5                 & \textbf{41.8}        & \textbf{42.9}        & 33.4                 & 29.7                 & \textbf{30.4}        & 21.7                 & \multicolumn{1}{c|}{22.6}          & \textbf{31.2}        \\ \hline
                                                          & \multicolumn{1}{l}{}        & \multicolumn{1}{l}{}        & \multicolumn{1}{l}{}        & \multicolumn{1}{l}{}        & \multicolumn{1}{l}{}        & \multicolumn{1}{l}{}        & \multicolumn{1}{l}{}        & \multicolumn{1}{l}{}        & \multicolumn{1}{l}{}        & \multicolumn{1}{l}{}        & \multicolumn{1}{l}{}        & \multicolumn{1}{l}{}        & \multicolumn{1}{l}{}        & \multicolumn{1}{l}{}        & \multicolumn{1}{l}{}                             & \multicolumn{1}{l}{}        \\ \hline
\multicolumn{17}{c}{\textbf{Probabilistic methods}}                                                                                                                                                                                                                                                                                                                                                                                                                                                                                                                             \\ \hline
\multicolumn{1}{l|}{\textbf{Protocol \#2: P-MPJPE}}       & Dir.                        & Disc.                       & Eat                         & Greet                       & Phone                       & Photo                       & Pose                        & Pur.                        & Sit                         & SitD.                       & Smoke                       & Wait                        & WalkD.                      & Walk                        & \multicolumn{1}{c|}{WalkT.}                      & \textbf{Avg.}               \\ \hline
\multicolumn{1}{l|}{MHFormer~\cite{li2022mhformer}($N$=351,$H$=3)}         & 31.5                 & 34.9                 & 32.8                 & 33.6                 & 35.3                 & 39.6                 & 32.0                 & 32.2                 & 43.5                 & 48.7                 & 36.4                 & 32.6                 & 34.3                 & 23.9                 & \multicolumn{1}{c|}{25.1}          & 34.4                 \\
\multicolumn{1}{l|}{GFPose~\cite{ci2023gfpose}($H$ = 10)}                      & 32.0                 & 39.5                 & 34.4                 & 34.7                 & 38.6                 & 44.3                 & 32.7                 & 31.9                 & 49.0                 & 60.1                 & 38.9                 & 36.6                 & 42.2                 & 28.3                 & \multicolumn{1}{c|}{32.3}          & 38.4                 \\
\multicolumn{1}{l|}{D3DP~\cite{shan2023diffusion}($N$=243, $*$)}              & 30.6                 & 32.4                 & 29.2                 & 30.9                 & 31.9                 & 37.4                 & 30.2                 & 29.3                 & \textbf{40.4}        & 43.2                 & \textbf{33.2}        & 30.4                 & 31.3                 & \textbf{21.5}        & \multicolumn{1}{c|}{22.3}          & 31.6                 \\ \hline 
\multicolumn{1}{l|}{Ours($N$=243,$H$=5, $W$=1)}                 & 29.5                 & \textbf{31.8}        & 28.7                 & \textbf{30.1}        & 31.3                 & 35.6                 & 29.7                 & 29.5                 & 41.7                 & 42.9                 & 33.4                 & \textbf{29.6}        & \textbf{30.4}        & 21.7                 & \multicolumn{1}{c|}{22.6}          & \textbf{31.2}        \\
\multicolumn{1}{l|}{Ours($N$=243, $H$=20, $W$=10)}              & \textbf{29.4}        & 32.1                 & \textbf{28.4}        & \textbf{30.1}        & \textbf{31.0}        & \textbf{35.3}        & \textbf{29.5}        & \textbf{29.2}        & 41.7                 & \textbf{42.7}        & \textbf{33.2}        & 29.7                 & 30.6                 & \textbf{21.5}        & \multicolumn{1}{c|}{\textbf{22.2}} & \textbf{31.2}        \\ \hline
\end{tabular}
\caption{Results on Human3.6M in millimeters under P-MPJPE. $N$, $H$, $W$: the number of input frames, hypotheses, and iterations used in the inference stage. In this table, we are compared with the deterministic and probabilistic methods. The best results are highlighted in bold. ($*$)-For clarity, $H$=20, $W$=10 is omitted. }
\end{table*}
\endgroup

\subsection{Implementation Details}
Our model is implemented in Pytorch ~\cite{paszke2019pytorch}, using AdamW ~\cite{loshchilov2017decoupled} as our optimizer with the momentum parameters as $\beta_{1}$, $\beta_{2}$ = 0.9, 0.999. We train our model for 400 epochs on NVIDIA Tesla A800 and the initial learning rate is 1$e^{-4}$, with a shrink factor of 0.993. We set batch size to 4 and in each batch, we use the length of 243 frames to train our model. We use a total of eight layers of spatio-temporal transformers in our model. The timestep $t$ is sampled from $U(0,\mathcal{T})$ during training, where $\mathcal{T}$ is the maximum number of timesteps and is set to 1000.
The first layer employed the spatio-temporal transformer in MixSTE ~\cite{Zhang_2022_CVPR}, followed by seven layers of HRST and HRTT used in a loop.

\subsection{Inference Details}
During the inference stage, we simultaneously sample $H$ bone length and bone direction hypotheses from a standard normal Gaussian distribution $\mathcal{N}(0,I)$ at the initial timestep $\mathcal{T}$ for the diffusion-based HSTDenoiser. The noisy 3D bone length $l_{0:H,\mathcal{T}}$ and bone direction $d_{0:H,\mathcal{T}}$, with the condition of the 2D pose sequence, are denoised by our denoiser and make predictions of 3D poses $\tilde{y}_{0:H,0}$. Then we can decompose $\tilde{y}_{0:H,0}$ to get the feasible predictions of bone length $\tilde{l}_{0:H,0}$ and bone direction $\tilde{d}_{0:H,0}$.
\par The denoised $\tilde{l}_{0:H,0}$ and $\tilde{d}_{0:H,0}$ can be used to generate the noisy input $l_{0:H,t^{'}}$ and $d_{0:H,t^{'}}$ at timestep $t^{'}$ via DDIM\cite{song2020denoising}, which is formulated in Eq (4) and shown in Figure 5. The updated input $l_{0:H,t^{'}}$ and $d_{0:H,t^{'}}$ will lead to another refined prediction and this procedure can be repeated for $W$ times to get the final predictions, which can enhance the prediction performance to some extent.

\begin{figure}[h]
\centering
\includegraphics[width=1\columnwidth]{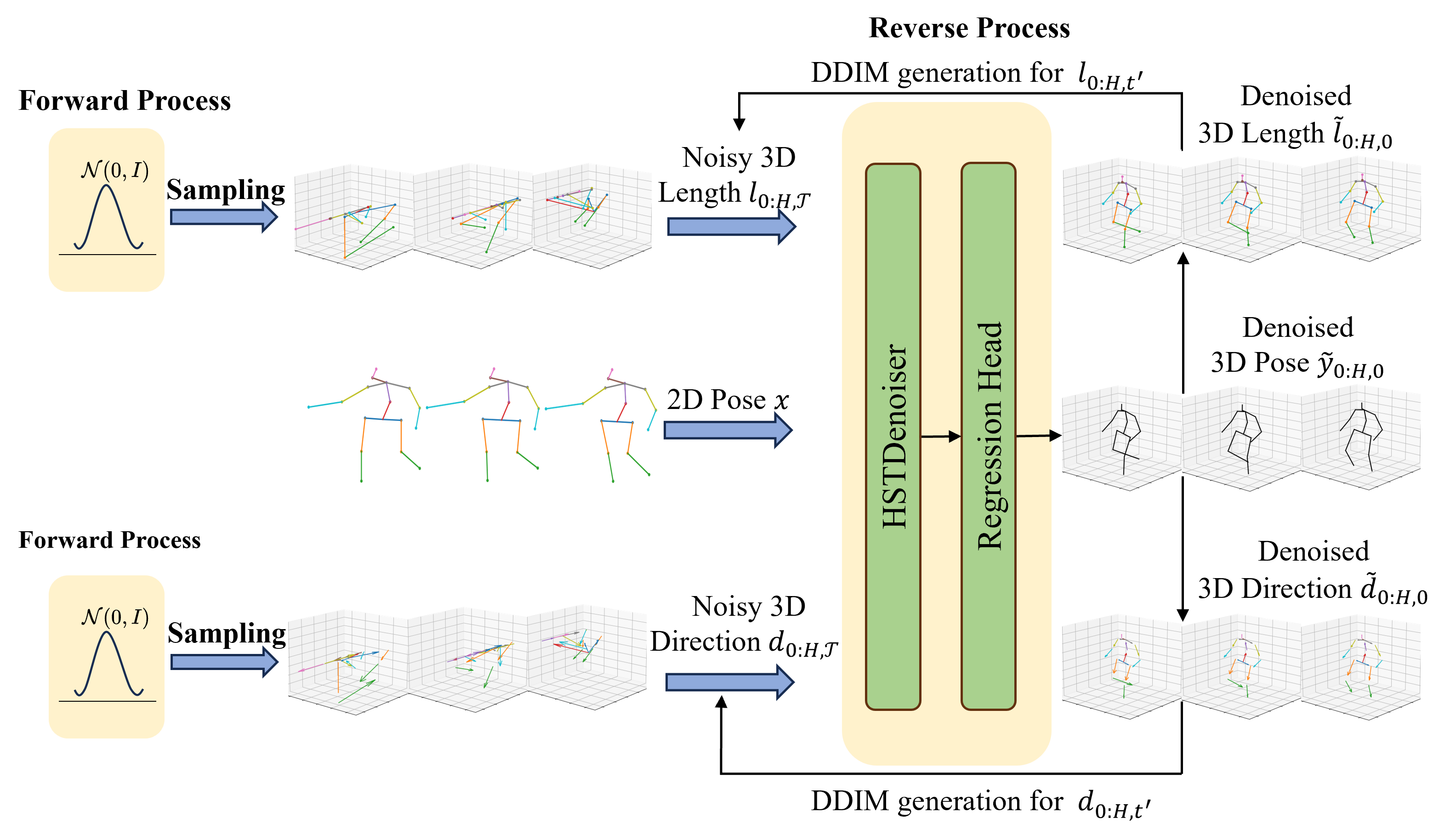} % Reduce the figure size so that it is slightly narrower than the column. Don't use precise values for figure width.This setup will avoid overfull boxes.
\caption{The overview of the inference pipeline.}
\label{fig4}
\end{figure}

\subsection{Additional Quantitative Results}
The supplement comparison of P-MPJPE is shown in Table 7. We compare our method with the SOTA deterministic methods and probabilistic methods. Based on whether the regression of the 3D pose locations is decomposed into the regression of bone length and bone direction, we divide the methods into disentangle-based methods and non-disentangle based methods. For disentangle-based methods, we can see from the table that our method achieves the best P-MPJPE of 31.2mm, which outperforms Anatomy3D~\cite{chen2021anatomy} by 3.7mm(10.6\%). While for non-disentangle based model, we improve STCFormer~\cite{tang20233d} by 0.6mm(1.9\%) under P-MPJPE. And then we compare our method with probabilistic methods, our method reaches the SOTA MPJPE of 31.2mm, outerperforms D3DP ~\cite{shan2023diffusion} by 0.4mm(1.3\%).

\subsection{Qualitative results}
We evaluate the qualitative results between our method, MixSTE~\cite{Zhang_2022_CVPR} and D3DP~\cite{shan2023diffusion}. The visual results are in Figure 6. Our method has better results than MixSTE and D3DP for the high hierarchical joints where the black dashed circle highlighting in the figure. 
\begin{figure}[h]
\centering
\includegraphics[width=1\columnwidth]{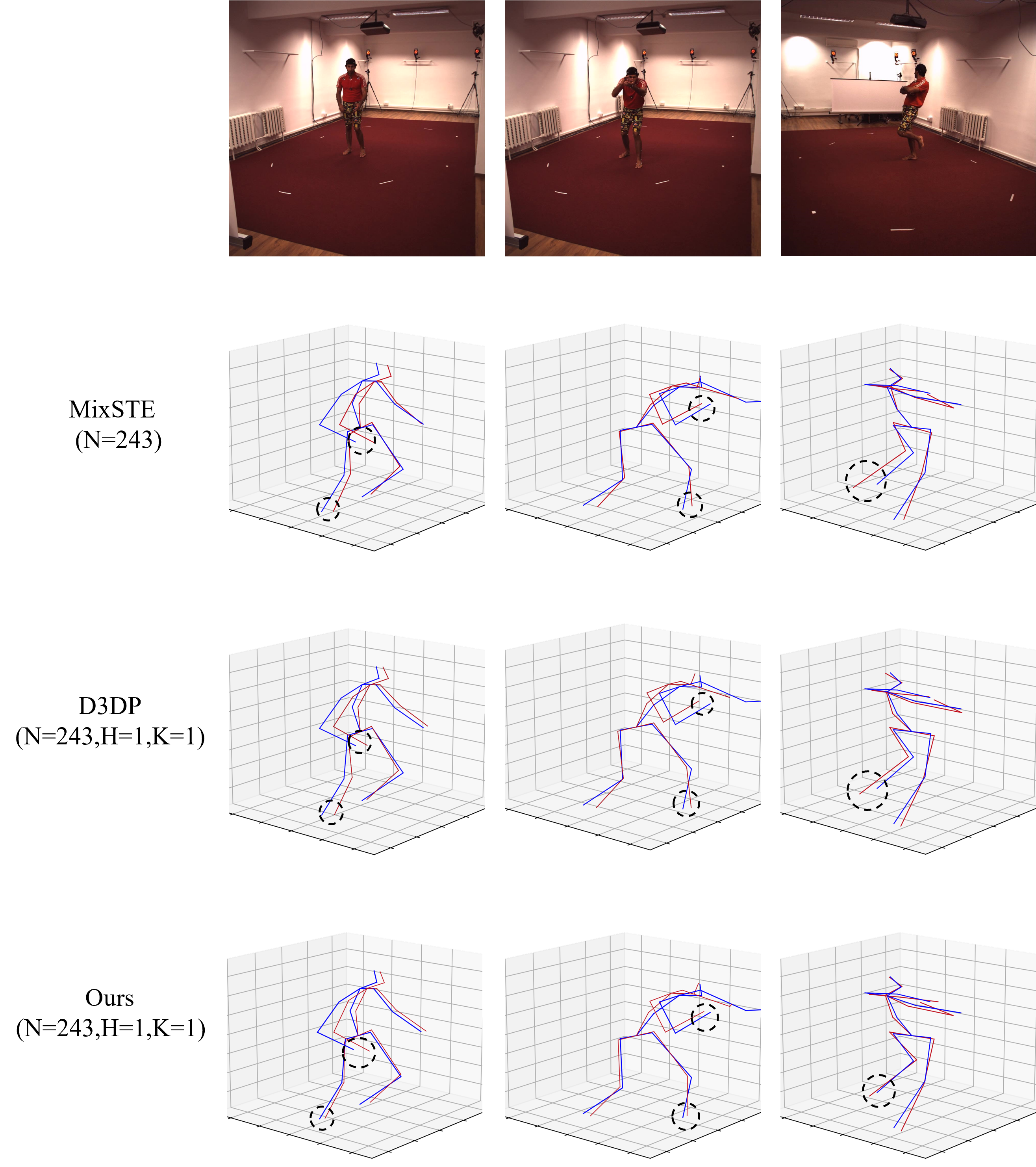} % Reduce the figure size so that it is slightly narrower than the column. Don't use precise values for figure width.This setup will avoid overfull boxes.
\caption{Qualitative comparison between our method, MixSTE and D3DP on Human3.6M. The ground truth pose is drawn in blue and the predicted pose is drawn in red. The black dashed circle highlights the locations where our method has better results.}
\label{fig4}
\end{figure}

\bibliography{aaai24}

\begin{thebibliography}{41}
\providecommand{\natexlab}[1]{#1}

\bibitem[{Baranchuk et~al.(2021)Baranchuk, Rubachev, Voynov, Khrulkov, and Babenko}]{baranchuk2021label}
Baranchuk, D.; Rubachev, I.; Voynov, A.; Khrulkov, V.; and Babenko, A. 2021.
\newblock Label-efficient semantic segmentation with diffusion models.
\newblock \emph{arXiv preprint arXiv:2112.03126}.

\bibitem[{Batzolis et~al.(2021)Batzolis, Stanczuk, Sch{\"o}nlieb, and Etmann}]{batzolis2021conditional}
Batzolis, G.; Stanczuk, J.; Sch{\"o}nlieb, C.-B.; and Etmann, C. 2021.
\newblock Conditional image generation with score-based diffusion models.
\newblock \emph{arXiv preprint arXiv:2111.13606}.

\bibitem[{Cao et~al.(2017)Cao, Simon, Wei, and Sheikh}]{cao2017realtime}
Cao, Z.; Simon, T.; Wei, S.-E.; and Sheikh, Y. 2017.
\newblock Realtime Multi-Person 2D Pose Estimation using Part Affinity Fields.
\newblock In \emph{CVPR}.

\bibitem[{Chen et~al.(2021)Chen, Fang, Shen, Zhu, Chen, and Luo}]{chen2021anatomy}
Chen, T.; Fang, C.; Shen, X.; Zhu, Y.; Chen, Z.; and Luo, J. 2021.
\newblock Anatomy-aware 3d human pose estimation with bone-based pose decomposition.
\newblock \emph{IEEE Transactions on Circuits and Systems for Video Technology}, 32(1): 198--209.

\bibitem[{Chen et~al.(2018)Chen, Wang, Peng, Zhang, Yu, and Sun}]{chen2018cascaded}
Chen, Y.; Wang, Z.; Peng, Y.; Zhang, Z.; Yu, G.; and Sun, J. 2018.
\newblock Cascaded pyramid network for multi-person pose estimation.
\newblock In \emph{Proceedings of the IEEE conference on computer vision and pattern recognition}, 7103--7112.

\bibitem[{Choi, Shim, and Kim(2022)}]{choi2022diffupose}
Choi, J.; Shim, D.; and Kim, H.~J. 2022.
\newblock Diffupose: Monocular 3d human pose estimation via denoising diffusion probabilistic model.
\newblock \emph{arXiv preprint arXiv:2212.02796}.

\bibitem[{Ci et~al.(2023)Ci, Wu, Zhu, Ma, Dong, Zhong, and Wang}]{ci2023gfpose}
Ci, H.; Wu, M.; Zhu, W.; Ma, X.; Dong, H.; Zhong, F.; and Wang, Y. 2023.
\newblock Gfpose: Learning 3d human pose prior with gradient fields.
\newblock In \emph{Proceedings of the IEEE/CVF Conference on Computer Vision and Pattern Recognition}, 4800--4810.

\bibitem[{Fan et~al.(2023)Fan, Chen, Chen, Cheng, Yuan, and Wang}]{fan2023frido}
Fan, W.-C.; Chen, Y.-C.; Chen, D.; Cheng, Y.; Yuan, L.; and Wang, Y.-C.~F. 2023.
\newblock Frido: Feature pyramid diffusion for complex scene image synthesis.
\newblock In \emph{Proceedings of the AAAI Conference on Artificial Intelligence}, volume~37, 579--587.

\bibitem[{Fang et~al.(2017)Fang, Xie, Tai, and Lu}]{fang2017rmpe}
Fang, H.-S.; Xie, S.; Tai, Y.-W.; and Lu, C. 2017.
\newblock {RMPE}: Regional Multi-person Pose Estimation.
\newblock In \emph{ICCV}.

\bibitem[{Gong et~al.(2023)Gong, Foo, Fan, Ke, Rahmani, and Liu}]{gong2023diffpose}
Gong, J.; Foo, L.~G.; Fan, Z.; Ke, Q.; Rahmani, H.; and Liu, J. 2023.
\newblock Diffpose: Toward more reliable 3d pose estimation.
\newblock In \emph{Proceedings of the IEEE/CVF Conference on Computer Vision and Pattern Recognition}, 13041--13051.

\bibitem[{Hagbi et~al.(2010)Hagbi, Bergig, El-Sana, and Billinghurst}]{hagbi2010shape}
Hagbi, N.; Bergig, O.; El-Sana, J.; and Billinghurst, M. 2010.
\newblock Shape recognition and pose estimation for mobile augmented reality.
\newblock \emph{IEEE transactions on visualization and computer graphics}, 17(10): 1369--1379.

\bibitem[{Ho, Jain, and Abbeel(2020)}]{ho2020denoising}
Ho, J.; Jain, A.; and Abbeel, P. 2020.
\newblock Denoising diffusion probabilistic models.
\newblock \emph{Advances in neural information processing systems}, 33: 6840--6851.

\bibitem[{Ho et~al.(2022)Ho, Saharia, Chan, Fleet, Norouzi, and Salimans}]{ho2022cascaded}
Ho, J.; Saharia, C.; Chan, W.; Fleet, D.~J.; Norouzi, M.; and Salimans, T. 2022.
\newblock Cascaded diffusion models for high fidelity image generation.
\newblock \emph{The Journal of Machine Learning Research}, 23(1): 2249--2281.

\bibitem[{Holmquist and Wandt(2022)}]{holmquist2022diffpose}
Holmquist, K.; and Wandt, B. 2022.
\newblock Diffpose: Multi-hypothesis human pose estimation using diffusion models.
\newblock \emph{arXiv preprint arXiv:2211.16487}.

\bibitem[{Ionescu et~al.(2014)Ionescu, Papava, Olaru, and Sminchisescu}]{h36m_pami}
Ionescu, C.; Papava, D.; Olaru, V.; and Sminchisescu, C. 2014.
\newblock Human3.6M: Large Scale Datasets and Predictive Methods for 3D Human Sensing in Natural Environments.
\newblock \emph{IEEE Transactions on Pattern Analysis and Machine Intelligence}, 36(7): 1325--1339.

\bibitem[{Kisacanin, Pavlovic, and Huang(2005)}]{kisacanin2005real}
Kisacanin, B.; Pavlovic, V.; and Huang, T.~S. 2005.
\newblock \emph{Real-time vision for human-computer interaction}.
\newblock Springer Science \& Business Media.

\bibitem[{Li et~al.(2023)Li, Shi, Dai, Zheng, Wang, Sun, Guo, Li, Zou, and Xiong}]{li2023pose}
Li, H.; Shi, B.; Dai, W.; Zheng, H.; Wang, B.; Sun, Y.; Guo, M.; Li, C.; Zou, J.; and Xiong, H. 2023.
\newblock Pose-Oriented Transformer with Uncertainty-Guided Refinement for 2D-to-3D Human Pose Estimation.
\newblock In \emph{Proceedings of the AAAI Conference on Artificial Intelligence}, volume~37, 1296--1304.

\bibitem[{Li et~al.(2022)Li, Liu, Tang, Wang, and Van~Gool}]{li2022mhformer}
Li, W.; Liu, H.; Tang, H.; Wang, P.; and Van~Gool, L. 2022.
\newblock Mhformer: Multi-hypothesis transformer for 3d human pose estimation.
\newblock In \emph{Proceedings of the IEEE/CVF Conference on Computer Vision and Pattern Recognition}, 13147--13156.

\bibitem[{Loshchilov and Hutter(2017)}]{loshchilov2017decoupled}
Loshchilov, I.; and Hutter, F. 2017.
\newblock Decoupled weight decay regularization.
\newblock \emph{arXiv preprint arXiv:1711.05101}.

\bibitem[{Mehta et~al.(2017)Mehta, Rhodin, Casas, Fua, Sotnychenko, Xu, and Theobalt}]{mono-3dhp2017}
Mehta, D.; Rhodin, H.; Casas, D.; Fua, P.; Sotnychenko, O.; Xu, W.; and Theobalt, C. 2017.
\newblock Monocular 3D Human Pose Estimation In The Wild Using Improved CNN Supervision.
\newblock In \emph{3D Vision (3DV), 2017 Fifth International Conference on}. IEEE.

\bibitem[{Nichol et~al.(2021)Nichol, Dhariwal, Ramesh, Shyam, Mishkin, McGrew, Sutskever, and Chen}]{nichol2021glide}
Nichol, A.; Dhariwal, P.; Ramesh, A.; Shyam, P.; Mishkin, P.; McGrew, B.; Sutskever, I.; and Chen, M. 2021.
\newblock Glide: Towards photorealistic image generation and editing with text-guided diffusion models.
\newblock \emph{arXiv preprint arXiv:2112.10741}.

\bibitem[{Paszke et~al.(2019)Paszke, Gross, Massa, Lerer, Bradbury, Chanan, Killeen, Lin, Gimelshein, Antiga et~al.}]{paszke2019pytorch}
Paszke, A.; Gross, S.; Massa, F.; Lerer, A.; Bradbury, J.; Chanan, G.; Killeen, T.; Lin, Z.; Gimelshein, N.; Antiga, L.; et~al. 2019.
\newblock Pytorch: An imperative style, high-performance deep learning library.
\newblock \emph{Advances in neural information processing systems}, 32.

\bibitem[{Pavlakos et~al.(2017)Pavlakos, Zhou, Derpanis, and Daniilidis}]{pavlakos2017coarse}
Pavlakos, G.; Zhou, X.; Derpanis, K.~G.; and Daniilidis, K. 2017.
\newblock Coarse-to-fine volumetric prediction for single-image 3D human pose.
\newblock In \emph{Proceedings of the IEEE conference on computer vision and pattern recognition}, 7025--7034.

\bibitem[{Pavllo et~al.(2019)Pavllo, Feichtenhofer, Grangier, and Auli}]{pavllo:videopose3d:2019}
Pavllo, D.; Feichtenhofer, C.; Grangier, D.; and Auli, M. 2019.
\newblock 3D human pose estimation in video with temporal convolutions and semi-supervised training.
\newblock In \emph{Conference on Computer Vision and Pattern Recognition (CVPR)}.

\bibitem[{Saharia et~al.(2022)Saharia, Ho, Chan, Salimans, Fleet, and Norouzi}]{saharia2022image}
Saharia, C.; Ho, J.; Chan, W.; Salimans, T.; Fleet, D.~J.; and Norouzi, M. 2022.
\newblock Image super-resolution via iterative refinement.
\newblock \emph{IEEE Transactions on Pattern Analysis and Machine Intelligence}, 45(4): 4713--4726.

\bibitem[{Shan et~al.(2022)Shan, Liu, Zhang, Wang, Ma, and Gao}]{shan2022p}
Shan, W.; Liu, Z.; Zhang, X.; Wang, S.; Ma, S.; and Gao, W. 2022.
\newblock P-stmo: Pre-trained spatial temporal many-to-one model for 3d human pose estimation.
\newblock In \emph{European Conference on Computer Vision}, 461--478. Springer.

\bibitem[{Shan et~al.(2023)Shan, Liu, Zhang, Wang, Han, Wang, Ma, and Gao}]{shan2023diffusion}
Shan, W.; Liu, Z.; Zhang, X.; Wang, Z.; Han, K.; Wang, S.; Ma, S.; and Gao, W. 2023.
\newblock Diffusion-Based 3D Human Pose Estimation with Multi-Hypothesis Aggregation.
\newblock \emph{arXiv preprint arXiv:2303.11579}.

\bibitem[{Sohl-Dickstein et~al.(2015)Sohl-Dickstein, Weiss, Maheswaranathan, and Ganguli}]{sohl2015deep}
Sohl-Dickstein, J.; Weiss, E.; Maheswaranathan, N.; and Ganguli, S. 2015.
\newblock Deep unsupervised learning using nonequilibrium thermodynamics.
\newblock In \emph{International conference on machine learning}, 2256--2265. PMLR.

\bibitem[{Song, Meng, and Ermon(2020)}]{song2020denoising}
Song, J.; Meng, C.; and Ermon, S. 2020.
\newblock Denoising diffusion implicit models.
\newblock \emph{arXiv preprint arXiv:2010.02502}.

\bibitem[{Song and Ermon(2020)}]{song2020improved}
Song, Y.; and Ermon, S. 2020.
\newblock Improved techniques for training score-based generative models.
\newblock \emph{Advances in neural information processing systems}, 33: 12438--12448.

\bibitem[{Sun et~al.(2018)Sun, Xiao, Wei, Liang, and Wei}]{sun2018integral}
Sun, X.; Xiao, B.; Wei, F.; Liang, S.; and Wei, Y. 2018.
\newblock Integral human pose regression.
\newblock In \emph{Proceedings of the European conference on computer vision (ECCV)}, 529--545.

\bibitem[{Tang et~al.(2023)Tang, Qiu, Hao, Hong, and Yao}]{tang20233d}
Tang, Z.; Qiu, Z.; Hao, Y.; Hong, R.; and Yao, T. 2023.
\newblock 3D Human Pose Estimation With Spatio-Temporal Criss-Cross Attention.
\newblock In \emph{Proceedings of the IEEE/CVF Conference on Computer Vision and Pattern Recognition}, 4790--4799.

\bibitem[{Tekin et~al.(2016)Tekin, Rozantsev, Lepetit, and Fua}]{tekin2016direct}
Tekin, B.; Rozantsev, A.; Lepetit, V.; and Fua, P. 2016.
\newblock Direct prediction of 3d body poses from motion compensated sequences.
\newblock In \emph{Proceedings of the IEEE Conference on Computer Vision and Pattern Recognition}, 991--1000.

\bibitem[{Vaswani et~al.(2017)Vaswani, Shazeer, Parmar, Uszkoreit, Jones, Gomez, Kaiser, and Polosukhin}]{Vaswani_Shazeer_Parmar_Uszkoreit_Jones_Gomez_Kaiser_Polosukhin_2017}
Vaswani, A.; Shazeer, N.; Parmar, N.; Uszkoreit, J.; Jones, L.; Gomez, A.; Kaiser, L.; and Polosukhin, I. 2017.
\newblock Attention is All you Need.
\newblock \emph{Neural Information Processing Systems,Neural Information Processing Systems}.

\bibitem[{Wang et~al.(2022)Wang, Zeng, Wang, Liu, and Wang}]{wang2022motion}
Wang, G.; Zeng, H.; Wang, Z.; Liu, Z.; and Wang, H. 2022.
\newblock Motion Projection Consistency Based 3D Human Pose Estimation with Virtual Bones from Monocular Videos.
\newblock \emph{IEEE Transactions on Cognitive and Developmental Systems}.

\bibitem[{Wang et~al.(2020)Wang, Sun, Cheng, Jiang, Deng, Zhao, Liu, Mu, Tan, Wang et~al.}]{wang2020deep}
Wang, J.; Sun, K.; Cheng, T.; Jiang, B.; Deng, C.; Zhao, Y.; Liu, D.; Mu, Y.; Tan, M.; Wang, X.; et~al. 2020.
\newblock Deep high-resolution representation learning for visual recognition.
\newblock \emph{IEEE transactions on pattern analysis and machine intelligence}, 43(10): 3349--3364.

\bibitem[{Xu et~al.(2020)Xu, Yu, Ni, Yang, Yang, and Zhang}]{xu2020deep}
Xu, J.; Yu, Z.; Ni, B.; Yang, J.; Yang, X.; and Zhang, W. 2020.
\newblock Deep kinematics analysis for monocular 3d human pose estimation.
\newblock In \emph{Proceedings of the IEEE/CVF Conference on computer vision and Pattern recognition}, 899--908.

\bibitem[{Zhang et~al.(2022{\natexlab{a}})Zhang, Tu, Yang, Chen, and Yuan}]{Zhang_2022_CVPR}
Zhang, J.; Tu, Z.; Yang, J.; Chen, Y.; and Yuan, J. 2022{\natexlab{a}}.
\newblock MixSTE: Seq2seq Mixed Spatio-Temporal Encoder for 3D Human Pose Estimation in Video.
\newblock In \emph{Proceedings of the IEEE/CVF Conference on Computer Vision and Pattern Recognition (CVPR)}, 13232--13242.

\bibitem[{Zhang et~al.(2022{\natexlab{b}})Zhang, Ye, Tu, Qin, Qin, Zhang, and Liu}]{zhang2022spatial}
Zhang, J.; Ye, G.; Tu, Z.; Qin, Y.; Qin, Q.; Zhang, J.; and Liu, J. 2022{\natexlab{b}}.
\newblock A spatial attentive and temporal dilated (SATD) GCN for skeleton-based action recognition.
\newblock \emph{CAAI Transactions on Intelligence Technology}, 7(1): 46--55.

\bibitem[{Zhao et~al.(2023)Zhao, Zheng, Liu, Wang, and Chen}]{Zhao_2023_CVPR}
Zhao, Q.; Zheng, C.; Liu, M.; Wang, P.; and Chen, C. 2023.
\newblock PoseFormerV2: Exploring Frequency Domain for Efficient and Robust 3D Human Pose Estimation.
\newblock In \emph{Proceedings of the IEEE/CVF Conference on Computer Vision and Pattern Recognition (CVPR)}, 8877--8886.

\bibitem[{Zheng et~al.(2021)Zheng, Zhu, Mendieta, Yang, Chen, and Ding}]{zheng20213d}
Zheng, C.; Zhu, S.; Mendieta, M.; Yang, T.; Chen, C.; and Ding, Z. 2021.
\newblock 3d human pose estimation with spatial and temporal transformers.
\newblock In \emph{Proceedings of the IEEE/CVF International Conference on Computer Vision}, 11656--11665.

\end{thebibliography}

\end{document}